%% file: main.tex
\documentclass[10pt,twocolumn,letterpaper]{article}

\usepackage[pagenumbers]{cvpr} 

\input{preamble}
\definecolor{cvprblue}{rgb}{0.21,0.49,0.74}
\usepackage[pagebackref,breaklinks,colorlinks,allcolors=cvprblue]{hyperref}

\usepackage{bm}
\usepackage{multirow} 
\usepackage{algpseudocode}
\usepackage{xcolor}
\usepackage{adjustbox}
\usepackage[ruled,vlined]{algorithm2e}
\usepackage{float}
\usepackage[accsupp]{axessibility}
\usepackage{hyperref}

\def\paperID{} 
\def\confName{CVPR}
\def\confYear{2026}

\title{SpiralDiff: Spiral Diffusion with LoRA for RGB-to-RAW Conversion \\ Across Cameras}

\author{
    Huanjing Yue$^{1,2}$ \enspace
    Shangbin Xie$^{1,2}$ \enspace
    Cong Cao$^1$ \enspace
    Qian Wu$^3$ \enspace
    Lei Zhang$^3$ \enspace
    Lei Zhao$^3$ \enspace
    Jingyu Yang$^1$\footnotemark[1] \\
    \textsuperscript{1}School of Electrical and Information Engineering, Tianjin University \\
    \textsuperscript{2}State key laboratory of Smart Power Distribution Equipment and System, Tianjin University \enspace
    \textsuperscript{3}Individual \\
    {\small \{huanjing.yue, chuancyx, caocong\_123, yjy\}@tju.edu.cn} \enspace
    {\small \{youzhagao, Zhaoleiyyy\}@gmail.com} \enspace
    {\small qian.wu.alex@outlook.com}
}

\begin{document}
\maketitle

\renewcommand{\thefootnote}{\fnsymbol{footnote}}
\footnotetext[1]{This work was supported in part by the National Natural Science
Foundation of China under Grant 62472308 and Grant 62231018. Corresponding author: Jingyu Yang}

\input{camera_ready/0_abstract}    
\input{camera_ready/1_intro}

\input{camera_ready/2_relatedwork}
\input{camera_ready/3_method}

\input{camera_ready/4_experiment}

\input{camera_ready/5_application}

\input{camera_ready/6_conclusion}
{
    \small
    \bibliographystyle{ieeenat_fullname}
    \bibliography{main}
}

\input{camera_ready/7_appendix}

\end{document}

%% file: camera_ready/0_abstract.tex
\begin{abstract}
RAW images preserve superior fidelity and rich scene information compared to RGB, making them essential for tasks in challenging imaging conditions. To alleviate the high cost of data collection, recent RGB-to-RAW conversion methods aim to synthesize RAW images from RGB. However, they overlook two key challenges: (i) the reconstruction difficulty varies with pixel intensity, and (ii) multi-camera conversion requires camera-specific adaptation. To address these issues, we propose SpiralDiff, a diffusion-based framework tailored for RGB-to-RAW conversion with a signal-dependent noise weighting strategy that adapts reconstruction fidelity across intensity levels. In addition, we introduce CamLoRA, a camera-aware lightweight adaptation module that enables a unified model to adapt to different camera-specific ISP characteristics.
Extensive experiments on four benchmark datasets demonstrate the superiority of SpiralDiff in RGB-to-RAW conversion quality and its downstream benefits in RAW-based object detection. Our code is available at \href{https://github.com/Chuancy-TJU/SpiralDiff}{https://github.com/Chuancy-TJU/SpiralDiff}.
\end{abstract}

%% file: camera_ready/1_intro.tex
\section{Introduction}
\label{sec:intro}

\begin{figure}[t]
  \centering
  \includegraphics[width=0.95\linewidth]{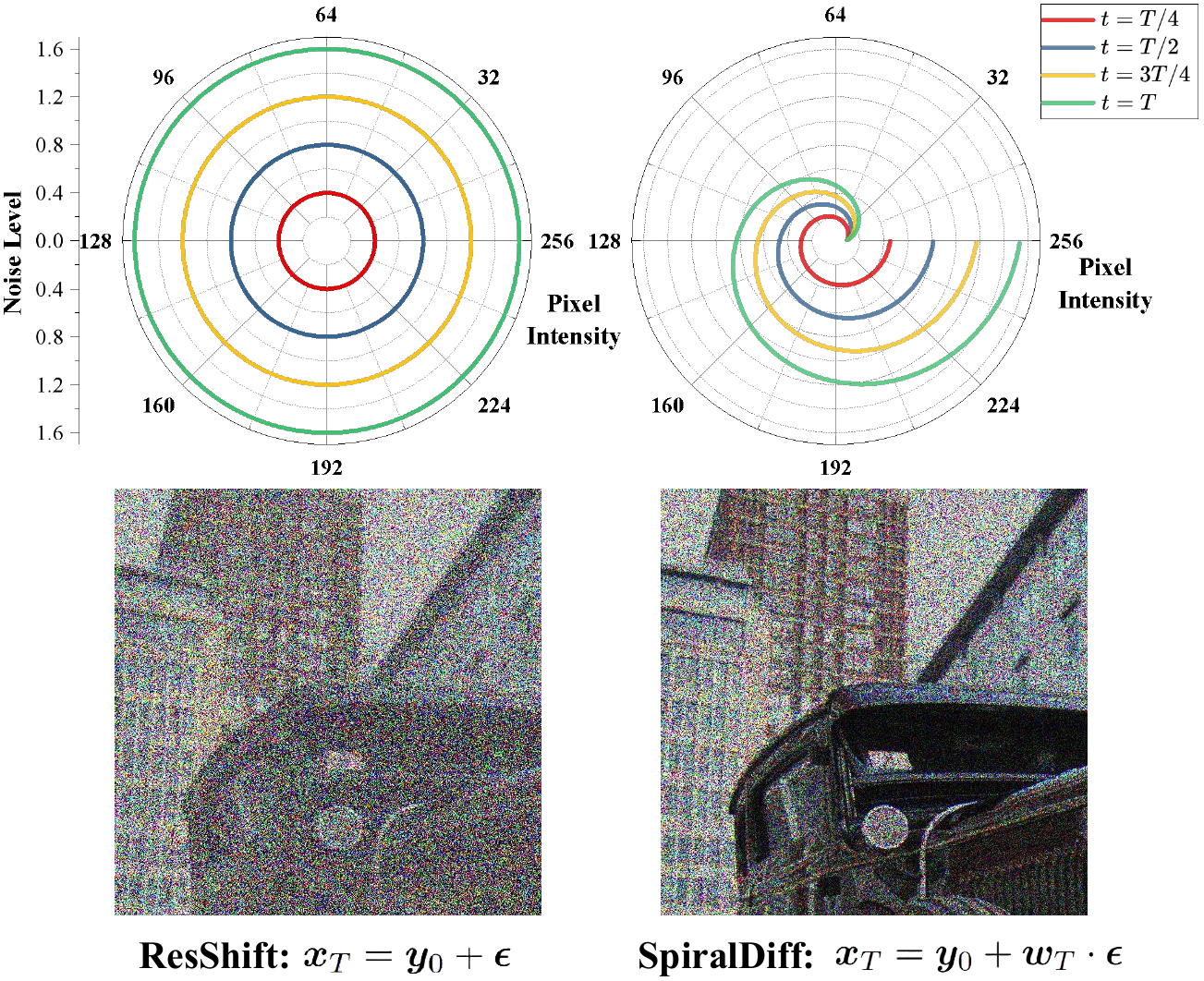}
   \caption{Comparison of noise schedule (top) and visualization of the noisy $\bm{x}_T$ (bottom) in diffusion: ResShift~\cite{yue2023resshift} uses uniform Gaussian noise, whereas our SpiralDiff introduces a signal-dependent noise schedule based on pixel intensity ($\bm{w}_t$).}
   \label{fig:figure1}
\end{figure}

As the direct output of a camera sensor, RAW images preserve rich, unprocessed scene information with a linear radiometric response and high dynamic range. These images are subsequently processed by an Image Signal Processor (ISP) to produce RGB images suitable for display. During this transformation, crucial photometric information is inevitably altered or discarded through operations such as demosaicing, white balance, tone mapping, and JPEG compression. As a result, the linearity, dynamic range, and realistic noise characteristics inherent in RAW data are largely lost in RGB images. 

Owing to these advantages, recent studies have demonstrated that performing computer vision tasks directly in the RAW domain can yield superior results in denoising~\cite{abdelhamed2018high,Yue_2020_CVPR}, low-light enhancement~\cite{chen2018learning,zhang2023towards}, and object detection~\cite{xu2023toward,li2025towards,Gamrian_2025_ICCV}.
Although existing RAW datasets provide valuable resources, they are limited in scale and diversity compared with RGB collections. Moreover, it is a costly and time-consuming process to construct a new RAW dataset for a specific sensor, often requiring extensive data collection and large storage resources. To alleviate this burden, recent studies have explored RGB-to-RAW conversion. It aims to invert the ISP pipeline and reconstruct plausible RAW images from abundantly available RGB inputs without repeatedly collecting new sensor-specific RAW datasets.

\begin{figure}[t]
  \centering
  \includegraphics[width=0.85\linewidth]{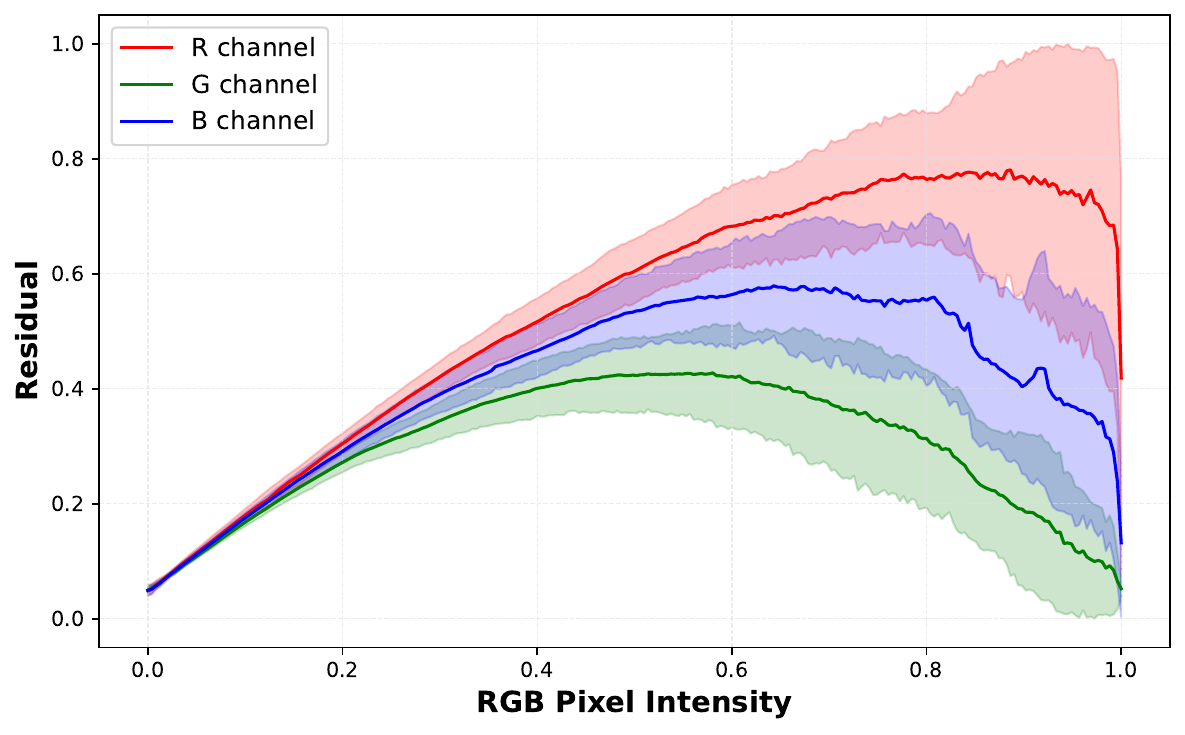}
   \caption{Relationship between RGB pixel intensity and residual magnitude between RGB and RAW images across channels on the FiveK Nikon dataset. Colored lines and shaded regions represent the mean and standard deviation, respectively.}
   \label{fig:residual}
\end{figure}

Existing RGB-to-RAW conversion methods can be broadly classified into metadata-based and metadata-free approaches. Metadata-based approaches~\cite{brooks2019unprocessing,li2023metadata,nam2022learning,punnappurath2021spatially,chen2024rawmamba} leverage ISP parameters or sampled RAW pixels, but such metadata is rarely accessible in real-world settings. 
Metadata-free methods~\cite{zamir2020cycleisp,xing2021invertible,conde2022model,berdan2025reraw,reinders2025raw} bypass this requirement by learning the mapping end-to-end. Although they achieve competitive results, they still overlook two fundamental challenges: 
(i) The conversion difficulty is inherently signal-dependent, varying with pixel intensity. 
As shown in Fig.~\ref{fig:residual}, residuals between RAW and RGB exhibit strong dependence on pixel intensity.
Specifically, in low-intensity regions, the residuals are small and stable, which enables high-fidelity recovery. In contrast, high-intensity and over-exposed regions suffer from large and uncertain residuals due to non-linear tone mapping and value clipping, making accurate prediction challenging. This indicates that a globally uniform reconstruction strategy in existing methods is suboptimal. Instead, an intensity-aware mechanism is required to adapt reconstruction flexibility according to local difficulty.
(ii) Effective multi-camera RGB-to-RAW conversion requires a unified model with camera-specific adaptation.
A naive approach to unification is to merge data from multiple cameras into a single training set. However, this leads to performance degradation because the model conflates different ISP characteristics and learns a compromised representation. This suggests a camera-specific adaptation is necessary to preserve the respective characteristics.

Based on the above analysis, we propose SpiralDiff, a diffusion-based framework tailored for RGB-to-RAW conversion. To handle intensity-dependent reconstruction, we introduce a spatially variant noise-weighting strategy within the diffusion process. Specifically, we build a Markov chain for the transition between the RAW image and its RGB counterpart based on ResShift~\cite{yue2023resshift, yue2024efficient}, an efficient diffusion framework for image restoration. 
We then modulate the isotropic Gaussian perturbation with a signal-dependent weight map: low-intensity regions receive relatively small perturbations to preserve fidelity, while high-intensity and over-exposed regions are injected with stronger noise to allow more flexible reconstruction under severe nonlinearity and clipping. An intuitive comparison is illustrated in Fig.~\ref{fig:figure1}.
The weighting strategy is further extended to a time-variant schedule that evolves with the diffusion steps, aligning the noise injection with the transformation process between RAW and RGB. In addition, to mitigate cross-camera interference when training a unified model, we propose CamLoRA, a camera-aware LoRA module. It consists of several lightweight LoRA layers that learn camera-specific adaptations conditioned on an input camera label. 
Such a design enables consistent camera-aware conversion within a unified model with minimal additional parameters.
Moreover, once the unified model is pretrained, it can be efficiently adapted to a new camera by fine-tuning only one LoRA layer, effectively leveraging the learned priors. Our major contributions are summarized as follows:

\begin{itemize}
\item We propose SpiralDiff, a diffusion-based framework tailored for RGB-to-RAW conversion. A signal-dependent noise weighting strategy is introduced to adaptively balance reconstruction fidelity and generative flexibility across regions with different intensity characteristics.

\item We introduce a camera-aware adaptation module (CamLoRA) for RGB-to-RAW conversion, enabling the model to generate RAW outputs that reflect device-specific properties conditioned on the input camera label.
\item Experiments demonstrate that SpiralDiff surpasses state-of-the-art approaches on four benchmark datasets. In addition, the generated RAW images improve downstream object detection performance.
\end{itemize}

%% file: camera_ready/2_relatedwork.tex
\section{Related Work}
\label{sec:Related}

\begin{figure*}[t]
  \centering
  \includegraphics[width=0.9\linewidth]{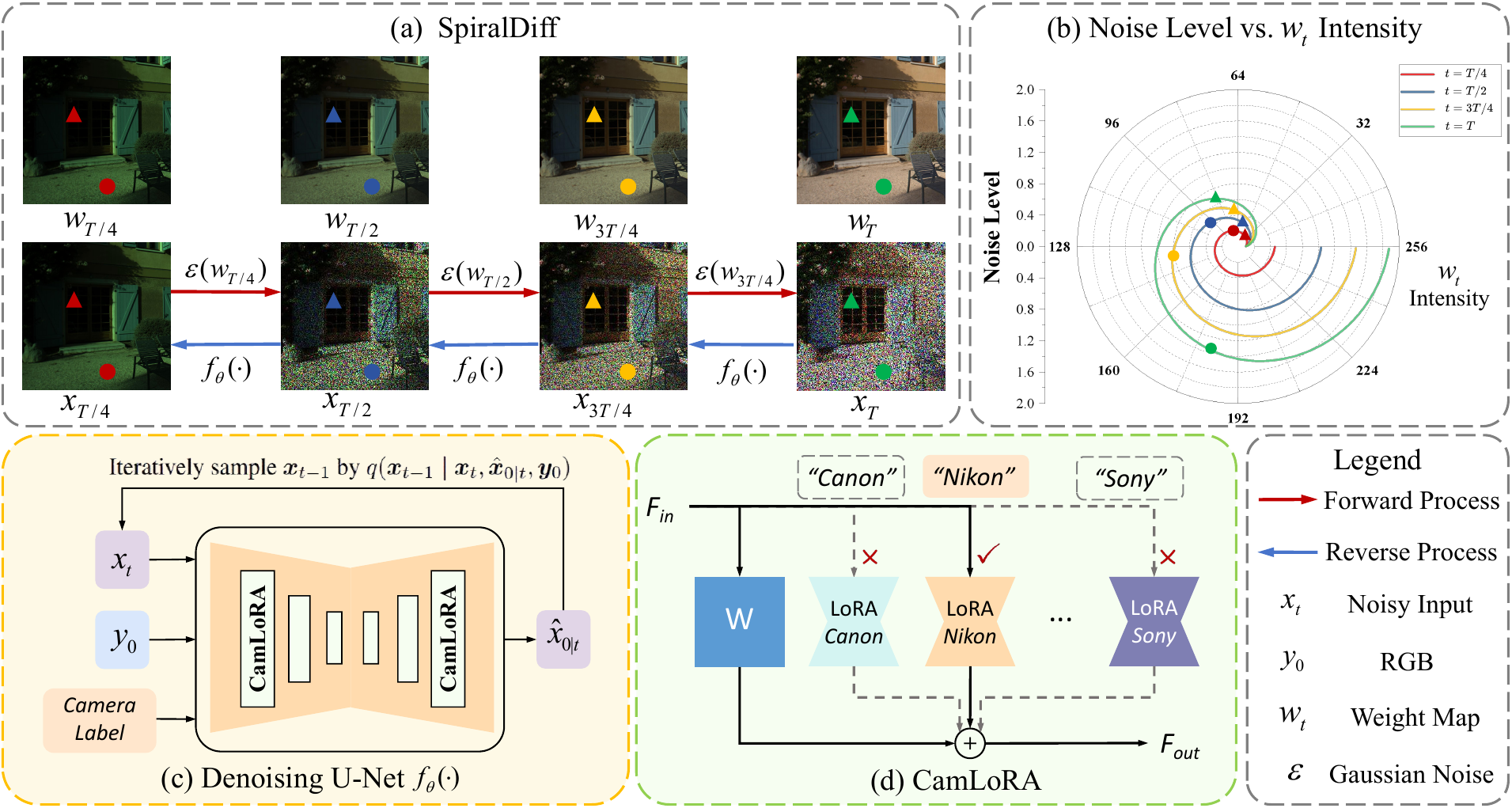}
   \caption{Overview of the proposed SpiralDiff with CamLoRA. 
   (a) SpiralDiff introduces a signal-dependent noise schedule via a weight map set $\{\bm{w}_t\}_{t=1}^T$ that aligns with the RAW-to-RGB conversion process.
   The noise level depends on local pixel intensity and diffusion timestep $t$: darker regions ($\triangle$) receive less noise, and brighter regions ($\bigcirc$) receive more. 
   (b) The spiral structure visualizes how noise scales with $\bm{w}_t$ for different pixel intensities. (c) shows the framework of SpiralDiff. The noisy image $\bm{x}_t$, RGB image $\bm{y}_0$ and camera label are fed into denoising U-Net, which iteratively samples to refine the RAW output. (d) The camera label selects the camera-specific LoRA layer in CamLoRA, enhancing adaptation to each camera's characteristic.}
   \label{fig:main}
\end{figure*}

\paragraph{RGB-to-RAW Conversion.}
The increasing need for large-scale datasets in the RAW domain has motivated research into synthesizing RAW images from RGB inputs. Existing approaches can be broadly divided into metadata-dependent and metadata-free methods. The former~\cite{brooks2019unprocessing,nam2022learning,chen2024rawmamba} reconstruct RAW images by leveraging ISP parameters or a subset of RAW data stored alongside the RGB image. However, such metadata is rarely available in real-world scenarios, limiting their practical applicability. Metadata-free RGB-to-RAW conversion aims to learn the inverse ISP mapping directly from data. Early works such as CycleISP~\cite{zamir2020cycleisp} and InvISP~\cite{xing2021invertible} enforce cycle consistency between RAW and RGB domains to regularize the ill-posed inversion. To make the model interpretable and parameters controllable, MBISPLD~\cite{conde2022model} replaces handcrafted ISP parameters in UPI~\cite{brooks2019unprocessing} with learnable dictionaries trained end-to-end. ReRAW~\cite{berdan2025reraw} employs a multi-head architecture to estimate RAW intensities. More recently, the generative approach RAW-Diffusion~\cite{reinders2025raw} formulates the task as a conditional generation process using diffusion models~\cite{ho2020denoising}. However, existing methods yield limited results, especially in high-intensity regions, because they treat all pixels uniformly and fail to account for the fact that reconstruction difficulty varies significantly with local intensity. To address this limitation, we propose SpiralDiff for intensity-aware RGB-to-RAW conversion.

\paragraph{Diffusion Models.}
Diffusion models~\cite{ho2020denoising} have gained significant attention in generative modeling due to their remarkable ability to synthesize high-fidelity images. These models operate through two complementary processes: a forward diffusion process that gradually adds Gaussian noise to data until it resembles a standard normal distribution, and a reverse denoising process that iteratively reconstructs the original signal using a neural network trained to predict and remove noise. Owing to their strong generative capacity and flexibility in conditioning, diffusion models have been successfully adapted to various image restoration tasks, including image super-resolution~\cite{wang2024exploiting,zhang2025uncertainty,wu2024seesr,xia2023diffir}, low-light image enhancement~\cite{hou2023global,jiang2023low,yi2023diff}, and high-dynamic-range imaging~\cite{hu2024generating,chen2025ultrafusion}. By conditioning on low-quality inputs, these methods are capable of generating high-quality outputs with rich structural and textural details. Among recent advances, ResShift~\cite{yue2023resshift} accelerates the sampling process requiring only four steps while achieving competitive performance across multiple image restoration tasks such as super-resolution, image inpainting, and blind face restoration. Given its high efficiency and strong performance on various restoration tasks, ResShift presents a promising foundation for RGB-to-RAW conversion.

\paragraph{Low-Rank Adaptation.}
As a parameter-efficient fine-tuning strategy, Low-Rank Adaptation (LoRA)~\cite{hu2022lora} decomposes weight updates into low-rank matrices, keeping the pretrained model frozen while optimizing only a small set of trainable parameters. Originally developed for large language models, LoRA has been successfully extended to computer vision tasks for domain adaptation~\cite{khanna2025explora,scheibenreif2024parameter} or stylized image generation~\cite{frenkel2024implicit,ouyang2025k,borse2024foura}. Recently, LoRA has shown promise in image restoration. UIR-LoRA~\cite{zhang2024uir} and LoRA-IR~\cite{ai2024lora} employ LoRA in universal image restoration as a lightweight alternative to degradation prompts, while PiSA-SR~\cite{sun2025pixel} uses it for controllable super-resolution.


%% file: camera_ready/3_method.tex
\section{Methodology}

Given an RGB image and its camera label, we aim to convert it into a RAW image using our proposed diffusion model. Our framework is presented in Fig.~\ref{fig:main}. The reverse process iteratively denoises a noisy RAW estimate, conditioned on the input RGB image and camera label with model weights dynamically modulated by CamLoRA. In the following, we first outline the baseline diffusion formulation, then detail SpiralDiff and CamLoRA.

\subsection{Preliminaries}
ResShift~\cite{yue2023resshift} is a diffusion model designed for image restoration. Instead of sampling from the pure Gaussian noise, it embeds the low-quality image into the initial state as the prior information, then progressively denoises and transitions to the high-quality image. Based on its residual shifting design, ResShift significantly improves efficiency requiring only four sampling steps. Considering its performance on various restoration tasks, it can be adopted for RGB-to-RAW conversion.

For ease of representation, we denote the RGB image by $\bm{y}_0$, the RAW image by $\bm{x}_0$ and their residual by $\bm{e}_0$, i.e., $\bm{e}_0=\bm{y}_0-\bm{x}_0$. ResShift builds up a transition from $\bm{x}_0$ to $\bm{y}_0$ by gradually shifting the residual $\bm{e}_0$. The forward transition is designed as
\begin{equation}\label{eq:reshift-forward}
q(\bm{x}_t | \bm{x}_{t-1}, \bm{y}_0) = \mathcal{N}(\bm{x}_t; \bm{x}_{t-1} + \alpha_t \bm{e}_0, \kappa^2 \alpha_t \bm{I}),
\end{equation}
for $t = 1,2,\dots,T$, where the shifting sequence $\{\eta_t\}_{t=1}^{T}$ is predefined and $\alpha_t=\eta_t-\eta_{t-1}$, $\kappa>0$ controls the overall noise level, and $\bm{I}$ is the identity matrix.
From this recursive definition, the marginal distribution at step $t$ can be derived as
\begin{equation}\label{eq:reshift-margin}
q(\bm{x}_t | \bm{x}_{0}, \bm{y}_0) = \mathcal{N}(\bm{x}_t; \bm{x}_{0} + \eta_t \bm{e}_0, \kappa^2 \eta_t \bm{I}).
\end{equation}
As $t$ increases, the mean of $\bm{x}_t$ is shifting gradually from $\bm{x}_0$ to $\bm{y}_0$, while the variance (i.e. the noise level) grows proportionally to $\eta_t$. Based on Eq.~\ref{eq:reshift-forward} and Eq.~\ref{eq:reshift-margin}, the corresponding reverse process is given by
\begin{equation}\label{eq:reshift-reverse}
    \begin{split}
    q(\bm{x}_{t-1}&|\bm{x}_t, \bm{x}_0, \bm{y}_0) \\
    &= \mathcal{N}( \bm{x}_{t-1};\frac{\eta_{t-1}}{\eta_t}\bm{x}_t + \frac{\alpha_t}{\eta_t}\bm{x}_0, \kappa^2 \frac{\eta_{t-1}}{\eta_t} \alpha_t \bm{I}).
    \end{split}
\end{equation}

\subsection{Spiral Diffusion}
\label{sec:spiral-method}
Although ResShift provides a strong foundation, it still faces challenges inherent to many existing methods: the difficulty of adapting reconstruction strategies across regions with varying pixel intensities. The reason is that commonly used isotropic Gaussian noise introduces uniform perturbations across all pixels, which fails to account for the fact that the reconstruction difficulty varies with local intensity characteristics.
Fig.~\ref{fig:residual} illustrates the relationship between RGB pixel intensity and the residual magnitude between RGB and RAW images across the red, green, and blue channels. The $x$-axis represents RGB pixel intensity, while the $y$-axis shows the mean residual magnitude for each channel. The red, green, and blue curves trace these mean values, and the shaded bands around each mean line represent the standard deviation above and below the mean. As pixel intensity increases, the bands widen, indicating that the residual variance is signal dependent. Note that, when the RGB intensities approach saturation, both the mean residual and its variance shrink, since the RAW image itself is also approaching saturation. The main reason is that multiplicative ISP operations, such as digital gains, cause brighter regions to exhibit larger residuals. For dark areas, excessive noise injection will hinder accurate reconstruction. Conversely, in bright or even over-exposed regions, insufficient noise injection limits the model's generation ability. This discrepancy makes it hard to balance reconstruction across different intensity regions with a fixed noise schedule. Based on these observations, we propose an adaptive signal-dependent noise weighting strategy in the forward process of the diffusion model, called SpiralDiff.

\paragraph{Forward Process.}
We introduce a weight map $\bm{w}$ to modulate the noise variance in Eq.\ref{eq:reshift-forward}. At timestep $T$, the local noise level of $\bm{x}_T$ is correlated with $\bm{y}_0$.
Low-intensity regions are perturbed lightly so the model can better recover their details, whereas high-intensity regions receive stronger noise to accommodate their higher reconstruction uncertainty. This design aligns with the analysis discussed above.

However, the static weight map does not adapt to the progress of the conversion. 
As $t \to 0$, we expect a stable output, where the weight map $\bm{w}_0$ should correlate to $\bm{x}_0$, rather than $\bm{y}_0$. To address this, we extend the static weight map $\bm{w}$ into a time-varying weight map set $\{\bm{w}_t\}_{t=1}^T$, defined as
\begin{equation}\label{eq:weight map}
\bm{w}_t = \bm{x}_{0} + \eta_t \bm{e}_0.
\end{equation}
The weight map $\bm{w}_t$ is consistent with the mean term and it can be interpreted as the noise-free intermediate state between $\bm{x}_0$ and $\bm{y}_0$, intuitively illustrated in Fig.~\ref{fig:main} (a). Consequently, the noise level evolves along with the signal transformation process, enabling a smooth transition from $\bm{x}_0$ to $\bm{y}_0$.
The transition process is then modeled as:
\begin{equation}\label{eq:spiral-forward}
\begin{split}
    q(&\bm{x}_t | \bm{x}_{t-1}, \bm{y}_0) \\
    &= \mathcal{N}(\bm{x}_t; \bm{x}_{t-1} + \alpha_t \bm{e}_0, \kappa^2 (\eta_t \bm{w}_t^2 - \eta_{t-1}\bm{w}_{t-1}^2) \bm{I}),
\end{split}
\end{equation}
where $\bm{w}_t^2$ represents the element-wise square of $w_t(p)$ for each pixel $p$.
By iteratively applying this transition, the marginal distribution at timestep $t$ can be expressed as
\begin{equation}\label{eq:spiral-margin}
q(\bm{x}_t | \bm{x}_{0}, \bm{y}_0) = \mathcal{N}(\bm{x}_t; \bm{x}_{0} + \eta_t \bm{e}_0, \kappa^2 \eta_t \bm{w}_t^2 \bm{I}).
\end{equation}
Thus, the noise level at timestep $t$ is spatially correlated with the weight map $\bm{w}_t$, which is dependent on the current transition state $\bm{x}_{0} + \eta_t \bm{e}_0$.

\paragraph{Reverse Process.}
Building upon the forward process, the corresponding backward transition distribution can be derived from Eq.~\ref{eq:spiral-forward} and Eq.~\ref{eq:spiral-margin}, denoted as:
\begin{equation}\label{eq:spiral-reverse-q}
   q(\bm{x}_{t-1} | \bm{x}_{t}, \bm{x}_{0}, \bm{y}_0) = \mathcal{N}(\bm{x}_{t-1} ; \bm{\mu}_{t-1},\bm{\Sigma}_{t-1}),
\end{equation}
where the mean $\bm{\mu}_{t-1}$ and variance $\bm{\Sigma}_{t-1}$ are given by
\begin{equation}\label{eq:spiral-reverse}
\begin{gathered}
    \bm{\mu}_{t-1} = \bm{\gamma}_t (\bm{x}_t - \alpha_t \bm{e_0}) + (1-\bm{\gamma}_t) (\bm{x}_0 + \eta_{t-1} \bm{e}_0) \\
    \bm{\Sigma}_{t-1} = \kappa^2\bm{\gamma}_t (\eta_t\bm{w}_t^2-\eta_{t-1}\bm{w}_{t-1}^2 )\bm{I},\\
\end{gathered}
\end{equation}
with a blending coefficient $\bm{\gamma}_t = \frac{\eta_{t-1}\bm{w}_{t-1}^2}{\eta_t\bm{w}_t^2}$.

Notably, this backward transition distribution remains structurally aligned with that of ResShift (Eq.~\ref{eq:reshift-reverse}). On the one hand, when the spatial noise weighting is disabled ($\bm{w}_t\equiv1$), the expressions of $\bm{\mu}_{t-1}$ and $\bm{\Sigma}_{t-1}$ reduce exactly to those of ResShift, and SpiralDiff degenerates to ResShift. On the other hand, even in the general case, the reverse transition distribution in SpiralDiff preserves the same fundamental guidance mechanism as ResShift. Specifically, in both frameworks, the mean $\bm{\mu}_{t-1}$ is expressed as a convex combination of two guided components: a \emph{noisy term}, representing the current corrupted state, and a \emph{clean term}, representing the target structure to be recovered. These components are blended under the control of a time-dependent balancing coefficient, which governs their relative influence at each sampling step. In ResShift (Eq.~\ref{eq:reshift-reverse}), the noisy term is simply $\bm{x}_t$ and the clean term is $\bm{x}_0$. The mean term gradually converges toward $\bm{x}_0$ as $t$ decreases, progressively refining the output image. In SpiralDiff, the noisy term corresponds to $\bm{x}_t - \alpha_t \bm{e}_0$, a backward-projected estimate of the intermediate state at step $t-1$ (derived from Eq.~\ref{eq:spiral-forward}). The clean term corresponds to $\bm{x}_0 + \eta_{t-1} \bm{e}_0$, which represents a forward-advanced approximation targeting the same state $\bm{x}_{t-1}$ (Eq.~\ref{eq:spiral-margin}). The interpolation of these two terms occurs naturally within the local neighborhood of $\bm{x}_{t-1}$, yielding a more stable sampling trajectory. Crucially, the balancing coefficient $\bm{\gamma}_t$ in SpiralDiff depends on the spatial weight map $\bm{w}_t$, enabling a pixel-adaptive blending between the noisy and clean components, in contrast to the scalar coefficient used in ResShift.

Following ResShift, a deep neural network $f_{\bm{\theta}}(\bm{x}_t, \bm{y}_0, t)$ with parameter $\bm{\theta}$ is optimized to predict the target $\bm{x}_0$. For the detailed derivation, overall pipeline, and loss function, please refer to the supplementary material.

\subsection{Camera-aware Low-Rank Adaptation}
RAW images exhibit strong camera-dependent characteristics. Training a single conversion model on mixed multi-camera data often leads to interference between devices, resulting in degraded reconstruction quality.
To address this, we propose Camera-aware Low-Rank Adaptation (CamLoRA), a lightweight adaptation module that conditions the denoising network on camera identity. As shown in Fig.~\ref{fig:main}, CamLoRA treats the input camera label as a discrete control signal, so that camera-specific adaptations are isolated in the low-rank branches, leaving the shared backbone free to learn universal features.

CamLoRA works by augmenting each trainable weight matrix $\mathbf{W} \in \mathbb{R}^{d \times k}$ in the backbone with a camera-specific low-rank update:
\begin{equation}
    \mathbf{W}_i = \mathbf{W} + \Delta\mathbf{W}_i = \mathbf{W} + \mathbf{B}_i \mathbf{A}_i,
\end{equation}
where $\mathbf{A}_i \in \mathbb{R}^{r \times k}$ and $\mathbf{B}_i \in \mathbb{R}^{d \times r}$ are low-rank matrices for camera $i$, and $r \ll \min(d, k)$. This design adds only $r(d + k)$ parameters per camera, ensuring high parameter efficiency. We apply CamLoRA to the query, key, value, and output projection matrices ($\mathbf{W}_q, \mathbf{W}_k, \mathbf{W}_v, \mathbf{W}_o$) in all Swin Transformer~\cite{liu2021swin} layers of our backbone. During training, the shared base weights $\mathbf{W}$ are updated using all data to learn general features. In contrast, only the adapter $\Delta\mathbf{W}_i$ corresponding to the current camera label $c_i$ is updated. Following standard LoRA practice~\cite{hu2022lora}, we initialize $\mathbf{B}_i$ to zero and $\mathbf{A}_i$ with random Gaussian noise to ensure stable optimization.

%% file: camera_ready/4_experiment.tex
\section{Experiments}
\label{sec:Experiments}

\begin{table*}[]
\caption{Performance comparison of different methods trained on separate and combined datasets, measured by PSNR($\uparrow$) and SSIM($\uparrow$). \textit{+CamLoRA} denotes that CamLoRA is applied to the combined training setting. Best results under each setting are highlighted in \textbf{bold}.}\label{tab:results}
\centering
\footnotesize
\renewcommand{\arraystretch}{1.0} 
\setlength{\tabcolsep}{3pt} 
\begin{tabular}{l|cccc|cccc}
\toprule
\multicolumn{1}{c|}{\multirow{2}{*}{Methods}} & \multicolumn{4}{c|}{Separate} & \multicolumn{4}{c}{Combined} \\
\multicolumn{1}{c|}{} & FiveK Canon & FiveK Nikon & NOD Nikon & NOD Sony & FiveK Canon & FiveK Nikon & NOD Nikon & NOD Sony \\ \midrule
CycleISP~\cite{zamir2020cycleisp} & 37.93 / 0.9913 & 40.18 / 0.9920 & 50.11 / 0.9985 & 46.57 / 0.9975 & 38.06 / 0.9911 & 38.86 / 0.9918 & 48.68 / 0.9984 & 46.12 / 0.9971 \\
InvISP~\cite{xing2021invertible} & 36.81 / 0.9814 & 34.30 / 0.9163 & 48.29 / 0.9954 & 44.76 / 0.9922 & 35.37 / 0.9828 & 36.60 / 0.9850 & 47.73 / 0.9948 & 44.43 / 0.9907 \\
ReRAW~\cite{berdan2025reraw} & 35.08 / 0.9608 & 33.09 / 0.9265 & 48.25 / 0.9906 & 45.50 / 0.9788 & 32.70 / 0.9684 & 33.24 / 0.9648 & 46.39 / 0.9715 & 45.51 / 0.9817 \\
RAW-Diffusion~\cite{reinders2025raw} & 39.96 / 0.9890 & 39.68 / 0.9866 & 50.52 / 0.9954 & 47.31 / 0.9908 & 40.63 / 0.9889 & 40.35 / 0.9916 & 49.60 / 0.9950 & 46.80 / 0.9897 \\ \midrule
SpiralDiff & \textbf{42.82 / 0.9936} & \textbf{41.72 / 0.9925} & \textbf{53.64 / 0.9990} & \textbf{50.46 / 0.9980} & 41.99 / 0.9931 & 43.19 / 0.9947 & 52.06 / 0.9987 & 49.52 / 0.9974 \\
+ CamLoRA & \textemdash & \textemdash & \textemdash & \textemdash & \textbf{42.46 / 0.9934} & \textbf{43.82 / 0.9950} & \textbf{52.62 / 0.9988} & \textbf{50.08 / 0.9977} \\ \bottomrule
\end{tabular}
\end{table*}

\begin{table}[]
\caption{Quantitative comparison on the over-exposed test set.}\label{tab:results-oe}
\centering
\footnotesize
\renewcommand{\arraystretch}{1.0} 
\setlength{\tabcolsep}{8pt} 
\begin{tabular}{l|cccc}
\toprule
\multicolumn{1}{c|}{\multirow{2}{*}{Methods}} 
& \multicolumn{2}{c}{FiveK Canon} 
& \multicolumn{2}{c}{NOD Nikon} \\
\multicolumn{1}{c|}{} 
& PSNR & SSIM  
& PSNR & SSIM \\ \midrule
CycleISP~\cite{zamir2020cycleisp}       & 27.75 & 0.9833 & 35.86 & 0.9957 \\
InvISP~\cite{xing2021invertible}         & 25.15 & 0.9699 & 36.66 & 0.9905 \\
ReRAW~\cite{berdan2025reraw}          & 26.36 & 0.9664 & 38.56 & 0.9691 \\
RAW-Diffusion~\cite{reinders2025raw}  & 30.60 & 0.9739 & 40.05 & 0.9926 \\
SpiralDiff     & \textbf{31.10} & \textbf{0.9856} & \textbf{40.79} & \textbf{0.9973} \\ \bottomrule
\end{tabular}
\end{table}

\subsection{Datasets}
We conduct experiments on two publicly available datasets: the MIT-Adobe FiveK Dataset (FiveK) ~\cite{bychkovsky2011learning} and the Night Object Detection Dataset (NOD) ~\cite{morawski2022genisp}. FiveK comprises diverse scenes captured under various lighting and exposure conditions. NOD is a large-scale benchmark for low-light object detection in the RAW domain, with annotated images captured at night. Following~\cite{reinders2025raw}, we select RAW images captured by Canon EOS 5D and Nikon D700 from the FiveK dataset containing 777 and 590 images, respectively, and randomly split them into train/test sets with an 85/15 ratio for each camera. For the NOD dataset, we use RAW images from Nikon D750 and Sony RX100 VII, resulting in 4.0k and 3.2k images, respectively, and adopt the official train-test split. All RAW files are processed with the \textit{rawpy} library to generate the corresponding RGB images. 
Unlike prior works~\cite{reinders2025raw,berdan2025reraw}, we use the as-shot white balance coefficients stored in the RAW metadata for ISP simulation instead of auto-estimated white balance, which provides a more faithful color rendering and is beneficial for camera-aware RGB-to-RAW conversion.

\subsection{Implementation Details}
Following ResShift, we adopt a U-Net architecture equipped with Swin Transformer~\cite{liu2021swin} blocks. Regarding VQGAN~\cite{esser2021taming}, which compresses images into a lower resolution to reduce the computational cost for efficient diffusion, we remove it since it is trained on RGB images and such a latent representation is ill-suited for RAW data. Therefore, SpiralDiff operates directly in pixel space. We retain the original diffusion hyperparameters from ResShift, including the noise scale $\kappa$, the shifting schedule $\{\eta_t\}_{t=0}^T$, and the total number of timesteps $T=4$, ensuring high inference efficiency. The CamLoRA rank is set to $r=8$ to maintain parameter efficiency, introducing only $2.7\%$ additional parameters ($1.05\,M$) across four camera-specific LoRA adapters.

Each RAW image is first normalized using its black level and white level, then linearly scaled to the range $[-1, 1]$. For the training stage, each RAW image is packed to four-channel \textit{RGGB} format, then randomly cropped to the resolution $256 \times 256$, with a batch size set to 8. Its RGB counterpart is cropped to $512 \times 512$ to maintain spatial alignment. We use the Adam optimizer~\cite{kingma2014adam} with cosine learning rate annealing~\cite{loshchilov2016sgdr}, starting from an initial learning rate of $1 \times 10^{-4}$ and decaying to $1 \times 10^{-5}$. During evaluation, images are inferred in the full resolution to avoid block artifacts. All experiments are implemented in PyTorch~\cite{paszke2019pytorch} and run on a single NVIDIA GeForce RTX 3090 GPU.

\subsection{Comparison with State-of-the-Art Methods}
We compare our method with four state-of-the-art methods for RGB-to-RAW conversion: CycleISP~\cite{zamir2020cycleisp}, InvISP~\cite{xing2021invertible}, ReRAW~\cite{berdan2025reraw}, and RAW-Diffusion~\cite{reinders2025raw}. For a fair comparison, we retrain and evaluate all methods using identical data splits. We adopt two settings: \textit{separate} and \textit{combined}. 
In the separate setting, a dedicated model is trained for each of the four cameras (FiveK Canon, FiveK Nikon, NOD Nikon, and NOD Sony). 
In the \textit{combined} setting, all camera data is merged into one training set to train a unified model, where CamLoRA uses camera labels and assigns different labels to FiveK Nikon and NOD Nikon since they correspond to different camera sensors. All the methods are trained with the same patch size, except for ReRAW~\cite{berdan2025reraw}, for which we keep its original global-context design. For InvISP~\cite{xing2021invertible}, we remove its white balance pre-processing on RAW images so that all methods operate on the same data, ensuring the fair comparison.

\begin{figure*}[t]
  \centering
  \includegraphics[width=0.9\linewidth]{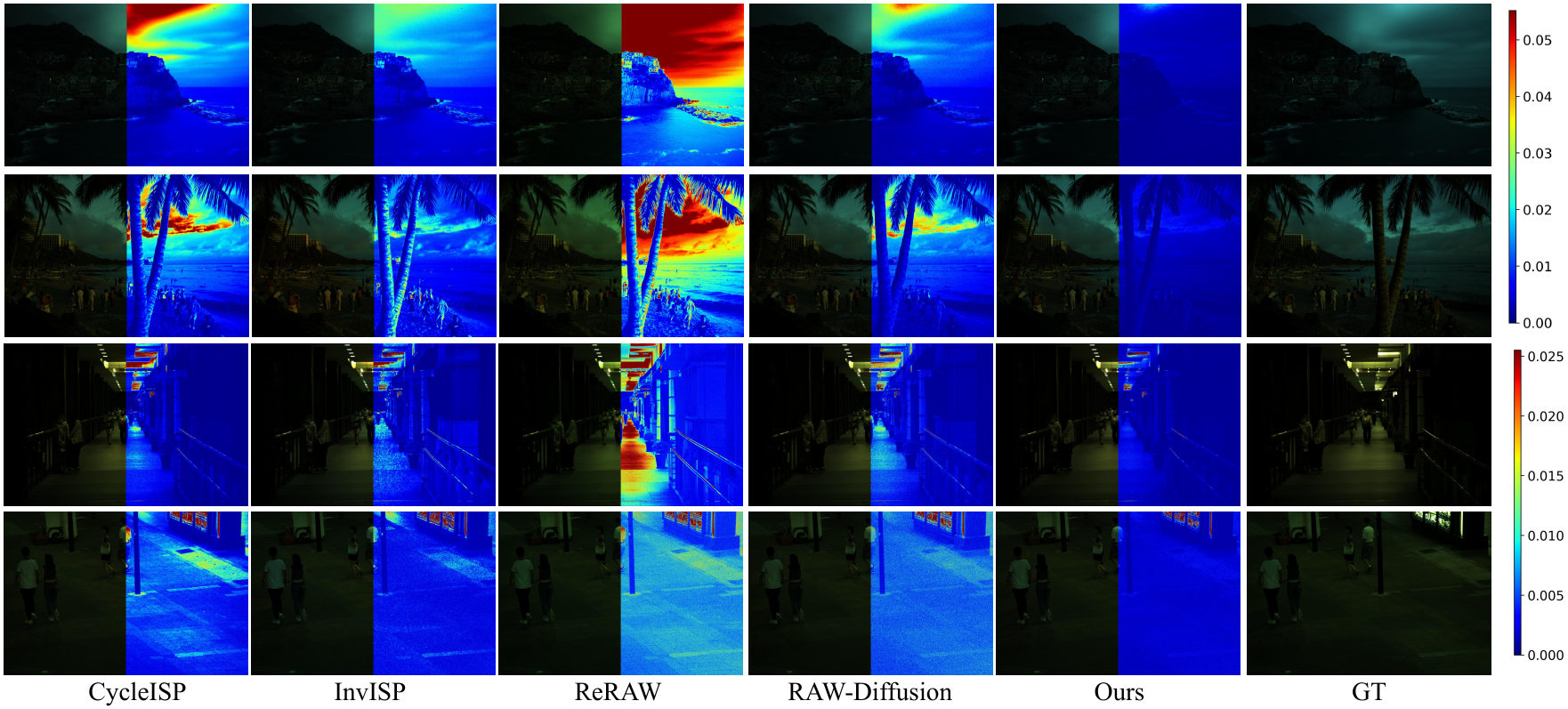}
   \caption{Qualitative comparison with state-of-the-art RGB-to-RAW conversion methods on the FiveK dataset (top two rows) and the NOD dataset (bottom two rows). For each result, the left half is the predicted RAW, and the right half is the error map. The proposed SpiralDiff shows better conversion results, especially in bright regions.}
   \label{fig:compare}
\end{figure*}

\paragraph{Quantitative Comparison.} 
Quantitative results on the four camera-specific test sets are reported in Tab.~\ref{tab:results}, under the \textit{Separate} and \textit{Combined} training settings mentioned before.
We evaluate conversion quality using PSNR and SSIM~\cite{wang2004image}.
In the separate setting, our proposed SpiralDiff achieves the best performance across all cameras, outperforming RAW-Diffusion~\cite{reinders2025raw} by a large margin. This highlights the advantage of our signal-dependent noise scheme over RAW-Diffusion’s use of isotropic Gaussian noise for modeling the RGB-to-RAW mapping. In the \textit{combined} setting, the model trained on the combined dataset usually performs worse than models trained on separate datasets, since one model needs to deal with multiple camera sensors. As shown in Tab.~\ref{tab:results}, all datasets exhibit performance degradation except for FiveK Nikon. This exception may be attributed to the fact that the separate model trained on FiveK Nikon uses only 502 training images, while the combined dataset includes a significantly larger number of Nikon-style images due to the incorporation of 3,206 NOD Nikon images. Fortunately, when equipped with our proposed CamLoRA, SpiralDiff achieves a clear improvement in the combined setting, denoted as \textit{+CamLoRA}. The results approach those of the separate setting. This demonstrates that our CamLoRA strategy can effectively adapt the model to different sensors according to the input camera label.

To further assess the generative performance of different methods, we construct an over-exposed (OE) test set by selecting images with higher proportions of saturated pixels. We select the top 20\% of images from the FiveK dataset and the top 10\% from the NOD dataset. The resulting OE test set includes 115 images: 24 from FiveK Canon, 18 from FiveK Nikon, 40 from NOD Nikon, and 33 from NOD Sony. As shown in Tab.~\ref{tab:results-oe}, the diffusion-based methods outperform previous methods, and our method outperforms the second best method (RAW-Diffusion) by more than 0.5 dB on PSNR. The full results are presented in the supplementary material. 

\paragraph{Qualitative Comparison.}
Fig.~\ref{fig:compare} shows qualitative comparisons of reconstructed RAW images by methods trained on the combined dataset. While all methods exhibit higher errors in bright regions, our method suffers least from color distortion and produces the most visually pleasing results. This demonstrates that our method outperforms others in preserving image quality and effectively handling high-intensity regions.

\subsection{Ablation Study}
First, we verify the effectiveness of the proposed noise weighting strategy by replacing it with other variants. All models are trained in the separate setting. The baseline variant is the original ResShift~\cite{yue2023resshift}, which utilizes isotropic Gaussian noise without adaptive weighting. The second variant directly uses the input RGB image ($\bm{y}_0$) to modulate the noise intensity. In contrast, our method uses the time-varying weight map sequence $\bm{w}_t$ to modulate the noise intensity. As shown in Tab.~\ref{tab:ablation-noise}, our method achieves the best results, outperforming the baseline method by more than 1 dB on FiveK Canon dataset. Note that an unsuitable noise weighting strategy, namely $\bm{y}_0$, performs even worse than the baseline, suggesting that a static weight map cannot model the reconstruction difficulty throughout the sampling process. This further demonstrates that our proposed noise weighting scheme is suitable for the RGB-to-RAW conversion task. Moreover, we conduct a plug-in experiment. We replace the DDPM diffusion process in RAW-Diffusion~\cite{reinders2025raw} with our SpiralDiff, while keeping its network architecture and training protocol unchanged, resulting in a +1.57 dB PSNR improvement on the FiveK Canon dataset (see supplementary for details).

\begin{table}[t]
\caption{Ablations on the noise weighting strategies. The baseline uses spatially uniform noise. $\bm{y}_0$ ($\bm{w}_t$) means the noise is weighted according to the intensity of $\bm{y}_0$ ($\bm{w}_t$).}\label{tab:ablation-noise}
\centering
\footnotesize
\renewcommand{\arraystretch}{1.0} 
\setlength{\tabcolsep}{5pt} 
\begin{tabular}{c|cccc}
\toprule
\multicolumn{1}{c|}{\multirow{2}{*}{Variant}} & \multicolumn{2}{c}{FiveK Canon} & \multicolumn{2}{c}{NOD Nikon} \\
\multicolumn{1}{c|}{} & PSNR & SSIM & PSNR & SSIM \\ \midrule
Baseline & 41.40 & 0.9906 & 53.48 & 0.9989 \\
$\bm{y}_0$ & 40.06 & 0.9912 & 53.42 & \textbf{0.9990} \\
$\bm{w}_t$ & \textbf{42.82} & \textbf{0.9936} & \textbf{53.64} & \textbf{0.9990} \\ \bottomrule
\end{tabular}
\end{table}

\begin{table}[t]
\caption{Ablations on three conditioning strategies: \textit{Uncond.} (no camera info), \textit{Embed.} (camera embedding), and \textit{CamLoRA}.}\label{tab:ablation-lora}
\centering
\footnotesize
\renewcommand{\arraystretch}{1.0} 
\setlength{\tabcolsep}{5pt} 
\begin{tabular}{c|cccc}
\toprule
\multicolumn{1}{c|}{\multirow{2}{*}{Variant}} & \multicolumn{2}{c}{FiveK Canon} & \multicolumn{2}{c}{NOD Nikon} \\
\multicolumn{1}{c|}{} & PSNR & SSIM & PSNR & SSIM \\ \midrule
Uncond. & 41.99 & 0.9931 & 52.06 & 0.9987 \\
Embed. & 40.97 & 0.9917 & 51.44 & 0.9984 \\
CamLoRA & \textbf{42.46} & \textbf{0.9934} & \textbf{52.62} & \textbf{0.9988}\\ 
\bottomrule
\end{tabular}
\end{table}

Second, we compare CamLoRA with a direct camera embedding approach~\cite{ho2022classifier}. It uses a set of trainable embeddings, one per camera, to provide camera-specific conditioning. Specifically, the embedding corresponding to the input camera label is injected at the same location as the time embedding to modulate features. We train a unified model with this approach, denoted as \textit{Embed}, and report results in Tab.~\ref{tab:ablation-lora}. Interestingly, it performs worse than even the unconditional baseline. 
In contrast, CamLoRA achieves substantial gains by applying low-rank updates, enabling effective camera adaptation with only a small number of parameters. In addition, we investigate the effect of CamLoRA rank: with only $r=4$, it already achieves a notable improvement over the unconditional baseline. Please refer to the supplementary material for detailed results.

CamLoRA enables the unified model to adapt to new cameras by training a new camera-specific LoRA branch while keeping shared parameters frozen, allowing the model to effectively reuse the rich priors learned during pre-training. This capability is particularly beneficial in few-shot settings. As shown in Tab.~\ref{tab:fewshot-lora}, with only 1 or 5 training images from the SID Sony dataset~\cite{chen2018learning}, CamLoRA fine-tuning achieves higher PSNR and SSIM than training a model from scratch, demonstrating superior adaptability and data efficiency. 

\subsection{Evaluation on Real-ISP Dataset}
\begin{figure}[t]
    \centering
    \includegraphics[width=0.9\linewidth]{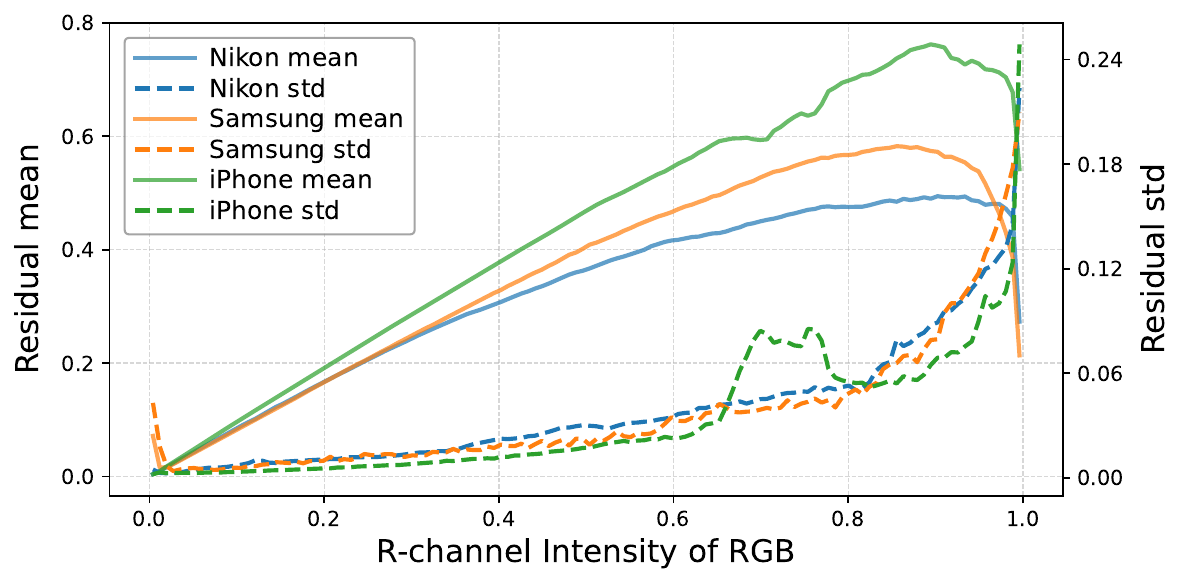}
    \caption{Residual-intensity relationship on rawpy ISP (FiveK Nikon) and real ISP (iPhone and Samsung).}
    \label{fig:redisual-rebuttal}
\end{figure}

We further evaluate SpiralDiff on a real-ISP dataset from NTIRE Challenge~\cite{conde2025raw}, where RGB images are rendered by smartphone ISPs (iPhone X and Samsung S9).
The residual-intensity relationship shows a similar increasing uncertainty trend to the rawpy-based setting (Fig.~\ref{fig:redisual-rebuttal}). Despite variations in the exact curve shapes across ISPs, this shared behavior is consistent and forms the basis of our design, which is therefore not tied to any specific ISP.
Replacing the original DDPM-based diffusion process in RAW-Diffusion with SpiralDiff while keeping the network architecture and training protocol unchanged yields consistent improvements on both subsets (Tab.~\ref{tab:ntire-plug-in}), showing that SpiralDiff remains effective on real-ISP data.

%% file: camera_ready/5_application.tex
\section{Applications}
A practical application of RGB-to-RAW conversion is object detection. Training detectors directly on RAW images can potentially exploit richer sensor information and skip the ISP processing to further reduce latency. With our method, existing large-scale RGB object detection datasets can be converted into the RAW domain with their annotations directly reused, avoiding the high cost of collecting and labeling RAW data. 
We evaluate whether synthesized RAW data improves RAW-domain detection on the NOD benchmark in low-data settings. As shown in Tab.~\ref{tab:object-detection}, training with only 100 real RAW images yields limited performance. Augmenting the training set with our synthesized RAW images from BDD100K (BDD)~\cite{yu2020bdd100k} and Cityscapes (CS)~\cite{cordts2016cityscapes} consistently improves the results. This highlights the value of leveraging large-scale RGB datasets and reduces the need for costly manual RAW annotation.

\begin{table}[t]
\caption{Few-shot RGB-to-RAW conversion on the SID Sony dataset. We compare training from scratch with CamLoRA adaptation under 1-shot and 5-shot settings.}
\label{tab:fewshot-lora}
\centering
\footnotesize
\renewcommand{\arraystretch}{1.0}
\setlength{\tabcolsep}{5pt}
\begin{tabular}{c|cccc}
\toprule
\multirow{2}{*}{Variant} 
    & \multicolumn{2}{c}{1-shot} 
    & \multicolumn{2}{c}{5-shot} \\
& PSNR & SSIM & PSNR & SSIM \\
\midrule
Scratch & 39.83 & 0.9895 & 43.40 & 0.9951 \\
CamLoRA & \textbf{42.85} & \textbf{0.9942} & \textbf{43.92} & \textbf{0.9954} \\
\bottomrule
\end{tabular}
\end{table}

\begin{table}[t]
\centering
\caption{Plug-in experiment on the real-ISP dataset.}
\label{tab:ntire-plug-in}
\setlength{\tabcolsep}{5pt}
\renewcommand{\arraystretch}{1.0}
\footnotesize
\begin{tabular}{l | c c}
\toprule
Method & iPhone & Samsung \\
\midrule
RAW-Diffusion (DDPM) & 22.74 / 0.7503 & 30.28 / 0.9140 \\
RAW-Diffusion (SpiralDiff) & \textbf{30.11} / \textbf{0.8810} & \textbf{34.07} / \textbf{0.9322} \\
\bottomrule
\end{tabular}
\end{table}

\begin{table}[t]
\caption{Comparison of object detection results on the NOD test set with different training datasets.}
\label{tab:object-detection}
\centering
\footnotesize
\renewcommand{\arraystretch}{1.0}
\setlength{\tabcolsep}{5pt}
\begin{tabular}{l|cccccc}
\toprule
\multirow{2}{*}{Dataset} 
    & \multicolumn{3}{c}{NOD Nikon} 
    & \multicolumn{3}{c}{NOD Sony} \\
& AP & AP\textsubscript{50} & AP\textsubscript{75} 
 & AP & AP\textsubscript{50} & AP\textsubscript{75} \\
\midrule
RGB-only        & 19.1 & 37.0 & 18.5 & 19.7 & 38.3 & 18.0 \\
RAW-only        & 18.4 & 35.1 & 17.0 & 17.6 & 35.7 & 15.1 \\
RAW + CS-RAW    & 24.5 & 46.2 & 23.5 & 25.9 & 49.3 & 24.5 \\
RAW + BDD-RAW   & \textbf{26.7} & \textbf{49.5} & \textbf{25.1} 
                & \textbf{29.0} & \textbf{55.1} & \textbf{26.5} \\
\bottomrule
\end{tabular}
\end{table}

%% file: camera_ready/6_conclusion.tex
\section{Conclusion}
In this work, we propose SpiralDiff, a diffusion-based framework tailored for RGB-to-RAW conversion. SpiralDiff incorporates a signal-dependent noise schedule that adapts reconstruction based on pixel intensity, allowing the model to handle both low-intensity and high-intensity regions effectively and in a spatially adaptive manner.
We further introduce CamLoRA, a parameter-efficient camera-aware adaptation module that enables a single model to adapt to multiple cameras. 
Experiments demonstrate that our method achieves state-of-the-art RGB-to-RAW conversion performance. 
In addition, the generated RAW images improve downstream object detection performance.

%% file: camera_ready/7_appendix.tex
\clearpage
\setcounter{section}{0}
\renewcommand\thesection{\Alph{section}}
\maketitlesupplementary

In this supplementary material, we present more implementation details of SpiralDiff and additional experiment results. First, we introduce the derivation of forward and backward processes and the overall training and sampling pipelines in Sec.~\ref{supp-diration}. Then we present additional experiment results in Sec.~\ref{supp-results}.

\section{Additional Details of SpiralDiff}
\label{supp-diration}
\paragraph{Derivation of Eq.~\ref{eq:spiral-margin}}
As discussed in Sec.~\ref{sec:spiral-method}, we propose a signal-dependent noise weighting strategy parameterized by a sequence of weight maps $\bm{w}_{t=1}^T$, defined as 
\begin{equation}\label{eq-supp:weight-map}
    \bm{w}_t = \bm{x}_{0} + \eta_t \bm{e}_0.
\end{equation}
Here, $\eta_t$ increases monotonically with $t$, with boundary conditions $\eta_0 = 0$ and $\eta_T \to 1$.
The forward transition distribution of SpiralDiff is given by:
\begin{equation}\label{eq-supp:forward-spiral}
q(\bm{x}_t \mid \bm{x}_{t-1}, \bm{x}_0, \bm{y}_0)
= \mathcal{N}\!\left(\bm{x}_t;\, \bm{x}_{t-1} + \alpha_t \bm{e}_0,\,
\kappa^2(\eta_t \bm{w}_t^2 - \eta_{t-1}\bm{w}_{t-1}^2)\bm{I}\right),
\end{equation}
where $\bm{w}_t^2$ denotes the element-wise square of $w_t^p$ for each pixel $p$ and $\bm{I}$ is the identity matrix. Following the reparameterization of the forward process, $\bm{x}_t$ can be expressed as
\begin{equation}\label{eq-supp:forward-reparameterization}
\begin{split}
\bm{x}_t &= \bm{x}_0 + \sum_{i=1}^{t} (\bm{x}_i - \bm{x}_{i-1}) \\
&= \bm{x}_0 + \sum_{i=1}^{t} \left( \alpha_i(\bm{y}_0 - \bm{x}_0)
    + \kappa \sqrt{\eta_i\bm{w}_i^2 - \eta_{i-1}\bm{w}_{i-1}^2}\, \bm{\epsilon}_i \right) \\
&= \bm{x}_0 + \eta_t (\bm{y}_0 - \bm{x}_0) + \kappa \sqrt{\eta_t}\, \bm{w}_t \bm{\epsilon}_t,
\end{split}
\end{equation}
where $\bm{\epsilon}_i \sim \mathcal{N}(\bm{0},\bm{I})$ are i.i.d. Gaussian noise variables, and $\bm{\epsilon}_t$ denotes the aggregated noise up to step $t$. This leads to the marginal distribution:
\begin{equation}\label{eq-supp:spiral-margin}
q(\bm{x}_t \mid \bm{x}_{0}, \bm{y}_0) = \mathcal{N}(\bm{x}_t; \bm{x}_{0} + \eta_t \bm{e}_0, \kappa^2 \eta_t \bm{w}_t^2 \bm{I}).
\end{equation}

\paragraph{Derivation of Eq.~\ref{eq:spiral-reverse-q}}
According to Bayes's theorem, the reverse transition distribution can be written as
\begin{equation}\label{eq-supp:bayes}
\begin{split}
q(\bm{x}_{t-1} \mid \bm{x}_t, \bm{x}_0, \bm{y}_0) &= \frac{q(\bm{x}_t \mid \bm{x}_{t-1}, \bm{x}_0, \bm{y}_0) q(\bm{x}_{t-1} \mid \bm{x}_0, \bm{y}_0)}{q(\bm{x}_t \mid \bm{x}_0, \bm{y}_0)} \\
&\propto q(\bm{x}_t \mid \bm{x}_{t-1}, \bm{x}_0, \bm{y}_0) q(\bm{x}_{t-1} \mid \bm{x}_0, \bm{y}_0).
\end{split}
\end{equation}
Incorporating Eq.\ref{eq-supp:forward-spiral}, Eq.\ref{eq-supp:spiral-margin} and Eq.\ref{eq-supp:bayes}, we consider the distribution at each pixel $p$ as:
\begin{equation}\label{eq-supp:a5}
q(x_{t-1}^p \mid x_t^p, x_0^p, y_0^p) \propto q(x_t^p \mid x_{t-1}^p, x_0^p, y_0^p) q(x_{t-1}^p \mid x_0^p, y_0^p)
\end{equation}
The two terms are:
\begin{equation}\label{eq-supp:a6-1}
q(x_t^p \mid x_{t-1}^p, x_0^p, y_0^p)
= \mathcal{N}\!\left(x_t^p;\, x_{t-1}^p + \alpha_t e_0^p,\,
\kappa^2 \left(\eta_t (w_t^p)^2 - \eta_{t-1} (w_{t-1}^p)^2\right)\right)
\end{equation}
\begin{equation}\label{eq-supp:a6-2}
q(x_{t-1}^p \mid x_0^p, y_0^p)
= \mathcal{N}\!\left(x_{t-1}^p;\, x_0^p + \eta_{t-1} e_0^p,\,
\kappa^2 \eta_{t-1} (w_{t-1}^p)^2\right)
\end{equation}
The exponent of the posterior $q(x_{t-1}^p \mid x_t^p, x_0^p, y_0^p)$ takes the following quadratic form:
\begin{equation}\label{eq-supp:a7}
\begin{aligned}
& -\frac{(x_t^p - x_{t-1}^p - \alpha_t e_0^p)^2}{2\kappa^2(\eta_t (w_t^p)^2 - \eta_{t-1}(w_{t-1}^p)^2)}
  - \frac{(x_{t-1}^p - x_0^p - \eta_{t-1} e_0^p)^2}{2\kappa^2 \eta_{t-1}(w_{t-1}^p)^2} \\
&= -\frac{1}{2}
   \left[
     \frac{1}{\kappa^2(\eta_t (w_t^p)^2 - \eta_{t-1}(w_{t-1}^p)^2)}
     + \frac{1}{\kappa^2 \eta_{t-1}(w_{t-1}^p)^2}
   \right](x_{t-1}^p)^2 \\
&\quad 
   + \left[
       \frac{x_t^p - \alpha_t e_0^p}{\kappa^2(\eta_t (w_t^p)^2 - \eta_{t-1}(w_{t-1}^p)^2)}
     + \frac{x_0^p + \eta_{t-1} e_0^p}{\kappa^2 \eta_{t-1}(w_{t-1}^p)^2}
     \right] x_{t-1}^p
     + \text{const} \\
&= -\frac{(x_{t-1}^p - \mu_{t-1}^p)^2}{2(\sigma_{t-1}^p)^2}
  + \text{const},
\end{aligned}
\end{equation}
where $\mu_{t-1}^p$ and $\sigma_{t-1}^p$ denote the mean and standard deviation of the posterior at pixel $p$, and $\mathrm{const}$ collects terms independent of $x_{t-1}^p$. The closed-form expressions for the posterior parameters are:
\begin{equation}\label{eq-supp:a8}
\begin{gathered}
    \mu_{t-1}^p = \gamma_t^p (x_t^p - \alpha_t e_0^p) + (1 - \gamma_t^p) (x_0^p + \eta_{t-1} e_0^p) \\
    (\sigma_{t-1}^p)^2 = \kappa^2\gamma_t^p \left( \eta_t (w_t^p)^2 - \eta_{t-1} (w_{t-1}^p)^2 \right) \\
    \gamma_t^p = \frac{\eta_{t-1} (w_{t-1}^p)^2}{\eta_t (w_t^p)^2}
\end{gathered}
\end{equation}
Therefore, the reverse transition distribution is given by:
\begin{equation}\label{eq-supp:a9}
\begin{split}    
    q(\bm{x}_{t-1} \mid \bm{x}_{t}, \bm{x}_{0}, \bm{y}_0) 
    &= \prod_{p} q(x^{p}_{t-1}\mid x^{p}_t, x^{p}_0, y^{p}_0) \\
    &= \mathcal{N}(\bm{x}_{t-1};\; \bm{\mu}_{t-1},\; \bm{\Sigma}_{t-1}),
\end{split}
\end{equation}
The mean and covariance are:
\begin{equation}\label{eq-supp:forward-spiral0}
\begin{gathered}
    \bm{\mu}_{t-1} = \bm{\gamma}_t \odot (\bm{x}_t - \alpha_t \bm{e}_0) + (\mathbf{1} - \bm{\gamma}_t) \odot (\bm{x}_0 + \eta_{t-1} \bm{e}_0), \\
    \bm{\Sigma}_{t-1} = \kappa^2 \, \bm{\gamma}_t \odot (\eta_t \bm{w}_t^2 - \eta_{t-1} \bm{w}_{t-1}^2), \\
    \bm{\gamma}_t = \frac{\eta_{t-1} \bm{w}_{t-1}^2}{\eta_t \bm{w}_t^2},
\end{gathered}
\end{equation}
where $\odot$ denotes element-wise multiplication and division is performed element-wise.

\paragraph{Details of $\bm{w}_t$}
Since $\bm{x}_0$ and $\bm{y}_0$ lie in the range $[-1,1]$, the resulting $\bm{w}_t$ also falls within this interval.
In practice, we normalize $\bm{w}_t$ to $[0,1]$ by a simple range mapping, and then add a small bias term $b \in (0,1)$:
\begin{equation}\label{eq-supp:w-hat}
    \hat{\bm{w}}_t = b + (1-b)\bm{w}_t
\end{equation}
This guarantees that $\hat{\bm{w}}_t$ is strictly bounded away from zero, where $b$ acts as a lower bound on the noise level and prevents numerical instability when computing $\bm{\gamma}_t$. Consequently, $\hat{\bm{w}}_t \in [b,1]$. We set $b=0.1$ as the default value.

\paragraph{Training and sampling pipelines}
We present the training and sampling pipelines in Alg.~\ref{alg:training} and Alg.~\ref{alg:sampling}.
During training, we randomly sample a triplet consisting of a RAW image $\mathbf{x}_0$, its corresponding RGB image $\mathbf{y}_0$, and the associated camera label $c$. Based on $\hat{\bm{w}}_t$ and $\bm{e}_0$, the noisy image $\bm{x}_t$ is sampled according to Eq.~\ref{eq-supp:spiral-margin}. The denoising model is optimized to predict $\bm{x}_0$, denoted as $\hat{\bm{x}}_{0\mid t} = f_\theta(\bm{x}_t, \bm{y}_0, t, c)$.
During sampling, $\hat{\bm{w}}_t$ and $\hat{\bm{e}}_0$ are updated according to the model output $\hat{\bm{x}}_{0\mid t}$, then $\bm{x}_{t-1}$ is sampled by Eq.~\ref{eq-supp:a9}. By iteratively denoising and sampling, the clean image $\bm{x}_0$ is obtained.

\begin{figure*}[t]
\centering
\begin{minipage}[t]{0.48\linewidth}
\begin{algorithm}[H]
\caption{Training}
\label{alg:training}
\KwIn{RAW image $\bm{x}_0$, RGB image $\bm{y}_0$, camera label $c$}
\While{not converged}{
    Sample $t \sim \mathcal{U}(\{1,\dots,T\})$\;
    Sample $\bm{\epsilon} \sim \mathcal{N}(\bm{0}, \bm{I})$\;
    $\bm{e}_0 = \bm{y}_0 - \bm{x}_0$\;
    $\bm{w}_t = \bm{x}_0 + \eta_t \bm{e}_0$\;
    $\hat{\bm{w}}_t = b + (1-b)\bm{w}_t$\;
    $\bm{x}_t = \bm{x}_{0} + \eta_t \bm{e}_0 + \kappa \sqrt{\eta_t} \hat{\bm{w}}_t \bm{\epsilon}$\;
    Take gradient step on $\nabla_\theta \mathcal{L}(\bm{x}_0, f_\theta(\bm{x}_t, \bm{y}_0, t, c))$\;
}
\Return{$\theta$}
\end{algorithm}
\end{minipage}
\hfill
\begin{minipage}[t]{0.48\linewidth}
\begin{algorithm}[H]
\caption{Sampling}
\label{alg:sampling}
\KwIn{RGB image $\bm{y}_0$, camera label $c$, denoising model $f_\theta$}

Sample $\bm{\epsilon} \sim \mathcal{N}(\bm{0}, \bm{I})$\;
$\hat{\bm{w}}_T = b + (1-b)\bm{y}_0$\;
$\bm{x}_T = \bm{y}_0 + \kappa \sqrt{\eta_T} \hat{\bm{w}}_T \bm{\epsilon}$\;

\For{$t = T, \dots, 1$}{
    Sample $\bm{\epsilon} \sim \mathcal{N}(\bm{0}, \bm{I})$\;
    $\hat{\bm{x}}_{0\mid t} = f_\theta(\bm{x}_t, \bm{y}_0, t, c)$\;
    $\hat{\bm{e}}_{0\mid t} = \bm{y}_0 - \hat{\bm{x}}_{0\mid t}$\;
    $\bm{w}_t = \hat{\bm{x}}_{0\mid t} + \eta_t \hat{\bm{e}}_{0\mid t}$\;
    $\bm{w}_{t-1} = \hat{\bm{x}}_{0\mid t} + \eta_{t-1}\hat{\bm{e}}_{0\mid t}$\;
    $\hat{\bm{w}}_t = b + (1-b)\bm{w}_t$\;
    $\hat{\bm{w}}_{t-1} = b + (1-b)\bm{w}_{t-1}$\;
    $\bm{\gamma}_t = \frac{\eta_{t-1}\hat{\bm{w}}_{t-1}^2}{\eta_t \hat{\bm{w}}_t^2}$\;
    
    $\bm{\mu}_{t-1} = \bm{\gamma}_t (\bm{x}_t - \alpha_t \hat{\bm{e}}_{0\mid t})
                     + (1-\bm{\gamma}_t)(\hat{\bm{x}}_{0\mid t} + \eta_{t-1} \hat{\bm{e}}_{0\mid t})$\;
    $\bm{\Sigma}_{t-1} = \kappa^2\bm{\gamma}_t(\eta_t\hat{\bm{w}}_t^2 - \eta_{t-1}\hat{\bm{w}}_{t-1}^2)\bm{I}$\;
    $\bm{x}_{t-1} = \bm{\mu}_{t-1} + \sqrt{\bm{\Sigma}_{t-1}} \odot \bm{\epsilon}$\;
}
\Return{$\bm{x}_0$}
\end{algorithm}
\end{minipage}
\end{figure*}

\paragraph{Loss function}
We follow the same loss function as that in RAW-Diffusion~\cite{reinders2025raw}, which combines the MSE, L1 and log-L1 loss. The overall loss is formulated as: 
\begin{equation}
\begin{split}
    \mathcal{L} &= \mathcal{L}_{\text{MSE}}+\mathcal{L}_{\text{L1}}+\mathcal{L}_{\text{logL1}} \\
    &= \sum_t \left( \| \hat{\bm{x}}_{0\mid t} - \bm{x}_0 \|_2^2 + \| \hat{\bm{x}}_{0\mid t} - \bm{x}_0 \|_1 + \| \log(\hat{\bm{x}}_{0\mid t}+ \epsilon) - \log(\bm{x}_0+ \epsilon) \|_1 \right),
\end{split}  
\end{equation}
where $\epsilon$ is a minimal constant.

\section{Additional Experiments and Results}
\label{supp-results}

This section provides additional quantitative evaluations. 

Tab.~\ref{tab:results-oe-full} reports the comparison across existing RGB-to-RAW conversion methods on the over-exposed test set, where SpiralDiff consistently outperforms previous approaches.

\begin{table}[H]
\caption{Quantitative comparison results on the over-exposed test set.}
\label{tab:results-oe-full}
\centering
\small
\renewcommand{\arraystretch}{1.0}
\setlength{\tabcolsep}{8pt}
\begin{tabular}{c|cccccccc}
\toprule
\multirow{2}{*}{Method} &
\multicolumn{2}{c}{FiveK Canon} &
\multicolumn{2}{c}{FiveK Nikon} &
\multicolumn{2}{c}{NOD Nikon} &
\multicolumn{2}{c}{NOD Sony} \\
& PSNR & SSIM & PSNR & SSIM & PSNR & SSIM & PSNR & SSIM \\
\midrule
CycleISP~\cite{zamir2020cycleisp}& 27.75 & 0.9833 & 30.19 & 0.9872 & 35.86 & 0.9957 & 32.52 & 0.9890 \\
InvISP~\cite{xing2021invertible} & 25.15 & 0.9699 & 29.11 & 0.9764 & 36.66 & 0.9905 & 30.85 & 0.9782 \\
ReRAW ~\cite{berdan2025reraw}    & 26.36 & 0.9664 & 27.51 & 0.9520 & 38.56 & 0.9691 & 34.96 & 0.9713 \\
RAW-Diffusion~\cite{reinders2025raw}  & 30.60 & 0.9739 & 32.95 & 0.9850 & 40.05 & 0.9926 & 35.58 & 0.9792 \\
SpiralDiff     & \textbf{31.10} & \textbf{0.9856} 
               & \textbf{34.42} & \textbf{0.9908} 
               & \textbf{40.79} & \textbf{0.9973} 
               & \textbf{36.11} & \textbf{0.9919} \\
\bottomrule
\end{tabular}
\end{table}

Tab.~\ref{tab:lora-rank} presents the ablation study on the LoRA rank $r$ used in CamLoRA. We observe that $r=8$ achieves the best performance and adopt it as the default setting in all experiments.

\begin{table}[H]
\caption{Ablation study on the LoRA rank $r$. We set $r=8$ as the default setting for all experiments.}
\label{tab:lora-rank}
\centering
\small
\renewcommand{\arraystretch}{1.0}
\setlength{\tabcolsep}{8pt}
\begin{tabular}{c|cccccccc}
\toprule
\multirow{2}{*}{Rank} &
\multicolumn{2}{c}{FiveK Canon} &
\multicolumn{2}{c}{FiveK Nikon} &
\multicolumn{2}{c}{NOD Nikon} &
\multicolumn{2}{c}{NOD Sony} \\
& PSNR & SSIM & PSNR & SSIM & PSNR & SSIM & PSNR & SSIM \\
\midrule
$r=0$   & 41.99 & 0.9931 & 43.19 & 0.9947 & 52.06 & 0.9987 & 49.52 & 0.9974 \\
$r=2$   & 42.64 & 0.9928 & 43.52 & 0.9948 & 52.23 & 0.9987 & 49.78 & 0.9977 \\
$r=4$   & 42.34 & 0.9931 & \textbf{43.95} & 0.9948 & 52.47 & 0.9988 & 49.82 & 0.9977 \\
$r=8$   & \textbf{42.46} & \textbf{0.9934} & 43.82 & \textbf{0.9950} & \textbf{52.62} & 0.9988 & \textbf{50.08} & 0.9977 \\
$r=16$  & 42.00 & 0.9932 & 42.82 & 0.9945 & 52.36 & \textbf{0.9989} & 49.60 & \textbf{0.9978} \\
\bottomrule
\end{tabular}
\end{table}

Tab.~\ref{tab:plugin} reports the plug-in experiment. RAW-Diffusion is built on the
DDPM~\cite{ho2020denoising}, and we replace this original diffusion process with
our SpiralDiff while leaving the network architecture and training settings
unchanged. The improvement, especially on the FiveK dataset with diverse
illumination conditions, validates the effectiveness of our signal-dependent
noise weighting strategy and shows that it can be integrated into other RGB-to-RAW reconstruction pipelines.



\begin{table}[H]
\caption{\textit{RAW-Diffusion (DDPM)} denotes the original RAW-Diffusion model with its DDPM-based diffusion process,
while \textit{RAW-Diffusion (SpiralDiff)} replaces only this diffusion process with our SpiralDiff formulation, keeping the network architecture unchanged.}
\label{tab:plugin}
\centering
\small
\renewcommand{\arraystretch}{1.2}
\setlength{\tabcolsep}{8pt}
\begin{tabular}{c|cccccccc}
\toprule
\multirow{2}{*}{Variant} &
\multicolumn{2}{c}{FiveK Canon} &
\multicolumn{2}{c}{FiveK Nikon} &
\multicolumn{2}{c}{NOD Nikon} &
\multicolumn{2}{c}{NOD Sony} \\
& PSNR & SSIM & PSNR & SSIM & PSNR & SSIM & PSNR & SSIM \\
\midrule
RAW-Diffusion (DDPM)     & 39.96 & 0.9890 & 39.68 & 0.9866 & 50.52 & \textbf{0.9954} & 47.31 & \textbf{0.9908} \\
RAW-Diffusion (SpiralDiff)  & \textbf{41.53} & \textbf{0.9912} 
                    & \textbf{42.34} & \textbf{0.9920} 
                    & \textbf{50.54} & \textbf{0.9954} 
                    & \textbf{47.32} & 0.9907 \\
\bottomrule
\end{tabular}
\end{table}

We conduct ablation experiments on a real-ISP dataset from the NTIRE'25 RGB-to-RAW conversion track~\cite{conde2025raw}, where RGB images are rendered by smartphone ISP pipelines (iPhone and Samsung). Since the official test set is not publicly available, we randomly split the provided training set into 85\%/15\% for training and evaluation. As shown in Tab.~\ref{tab:ntire}, SpiralDiff improves performance on both subsets, and further gains are obtained when integrating CamLoRA.

\begin{table}[H]
\centering
\begin{minipage}[t]{0.62\linewidth}
\centering
\caption{Results on the real-ISP dataset. Models are trained in the \textit{combined} setting.}
\label{tab:ntire}
\small
\renewcommand{\arraystretch}{1.2}
\setlength{\tabcolsep}{8pt}
\begin{tabular}{l|cccc}
\toprule
\multirow{2}{*}{Method} &
\multicolumn{2}{c}{iPhone} &
\multicolumn{2}{c}{Samsung} \\
& PSNR & SSIM & PSNR & SSIM \\
\midrule
Baseline & 30.79 & 0.9040 & 34.84 & 0.9362 \\
SpiralDiff & 31.56 & 0.9094 & 35.22 & 0.9394 \\
SpiralDiff + CamLoRA & \textbf{31.57} & \textbf{0.9100} & \textbf{35.96} & \textbf{0.9433} \\
\bottomrule
\end{tabular}
\end{minipage}\hfill
\begin{minipage}[t]{0.34\linewidth}
\centering
\caption{Object detection on NOD.}
\label{tab:od_nod}
\small
\renewcommand{\arraystretch}{1.2}
\setlength{\tabcolsep}{8pt}
\begin{tabular}{c c c c}
\toprule
Setting & \#Real & \#Syn & AP \\
\midrule
(a) & 1000 & 0   & 33.7 \\
(b) & 1000 & 732 & 35.9 \\
(c) & 1732 & 0   & 36.2 \\
\bottomrule
\end{tabular}
\end{minipage}
\end{table}

We further conduct object detection experiments in a data-rich setting. Tab.~\ref{tab:od_nod} reports three training settings: (a) 1000 real RAW images from NOD‑train, (b) the same 1000 real images augmented with 732 synthetic RAW images converted from NOD-val RGB using SpiralDiff, and (c) 1732 real RAW images from NOD-train+val. Setting (b) outperforms (a) and approaches (c), indicating that our synthetic RAW data serves as a low‑cost substitute for augmenting real RAW data in object detection. We also observe a trade-off when using cross-source synthetic data. When training with all real NOD-RAW together with synthesized CS-RAW, performance decreases on the NOD dataset but increases on the BDD dataset. This also happens when performing the same augmentation in the RGB domain. 
Overall, in data‑rich scenarios, same‑source synthetic data improves in‑dataset performance, whereas cross‑source synthetic data enhances generalization at the cost of degraded in‑dataset performance.

\clearpage